\documentclass[conference]{IEEEtran}
\usepackage{times}

\IEEEoverridecommandlockouts

\usepackage{graphics} 
\usepackage{epsfig} 

\usepackage{amsmath} 
\usepackage{amssymb}  
\usepackage{algorithm}
\usepackage[noend]{algpseudocode}
\usepackage[noadjust]{cite}
\usepackage{balance}

\newcounter{MYtempeqncnt}

\DeclareMathOperator*{\argmin}{argmin}

\begin{document}
\title{Model-Based Control Using Koopman Operators}

\author{\IEEEauthorblockN{Ian Abraham, Gerardo De La Torre\thanks{This work was supported by Army Research Office grant W911NF-14-1-0461. }, and
Todd D. Murphey}
\IEEEauthorblockA{Department of Mechanical Engineering\\
Northwestern University,
Evanston, Illinois 60208\\ Email:
 i-abr@u.northwestern.edu\\ gerardo.delatorre@northwestern.edu, \\t-murphey@northwestern.edu  }}

\maketitle

\begin{abstract}
This paper explores the application of Koopman operator theory to the control of robotic systems. The operator is introduced as a method to generate data-driven models that have utility for model-based control methods. We then motivate the use of the Koopman operator towards augmenting model-based control. Specifically, we illustrate how the operator can be used to obtain a linearizable data-driven model for an unknown dynamical process that is useful for model-based control synthesis. Simulated results show that with increasing complexity in the choice of the basis functions, a closed-loop controller is able to invert and stabilize a cart- and VTOL-pendulum systems. Furthermore, the specification of the basis function are shown to be of importance when generating a Koopman operator for specific robotic systems. Experimental results with the Sphero SPRK robot explore the utility of the Koopman operator in a reduced state representation setting where increased complexity in the basis function improve open- and closed-loop controller performance in various terrains, including sand.
\end{abstract}

\section{Introduction}
Modeling for complex dynamical systems has typically been the first step when designing, control, planning, or state-estimation algorithms. System design and specifications have been dependent on the use of high-fidelity models. However, any derivation of a dynamical model from first principles is typically a demanding task when the complexity of state interactions is high. Moreover, analytical models do not capture external disturbances. As a result, derived models, for use in model-based control settings, often have limited use or poor prediction over longer time spans. Nevertheless, a representation of the behavior of a dynamical system is central to most model-based engineering and scientific application.

Within the field of systems and control theory, model uncertainty has typically been mitigated with the use of robust and adaptive control architectures. Typically, adaptive controllers are self tuning and reactive to incoming state information while robust controllers are designed to be invariant to model uncertainty \cite{adaptive1,adaptive2,robust1,robust2}. Motion planning for uncertain dynamical systems have also been extensively investigated. Generally, in this approach, uncertainty is explicitly modeled and incorporated into the decision making process \cite{motion1,motion2,motion3}. However, like robust and adaptive control approaches, the need for an explicit uncertainty model often limits its utility in general settings. Machine learning, offers a much more general approach \cite{ml1,ml2,ml3}. In particular, recent advances have utilized large sets of data to perform model-based control of various dynamical systems \cite{williamsinformation}. Nonetheless, several questions about the training data, stability, convergence properties, computational complexity, and mechanical property conservation of the models are still open questions that need to be addressed.

Recently, the use of data-driven techniques to mitigate the effects of model uncertainty have sparked interest in the Koopman operator \cite{Koop1}. The Koopman operator is a infinite-dimensional linear operator that is able to exactly capture the behavior of nonlinear dynamical systems. In application, the Koopman operator is approximated with a finite-dimensional linear operator \cite{Koop2}. This approximation can be computed in a solely data-driven manner without any prior information of the dynamical system. Complex fluid flow systems have accurately been modeled using this approach \cite{Koop3}. Furthermore, it has been shown that the spectral properties of the approximate Koopman operator can be examined to investigate system-level behavior like ergodicity and stability \cite{Koop1, Koop4, Koop5}. In addition, recent work has shown its utility in human-machine systems \cite{broadRSS}. In this paper, we investigate the utility of Koopman operator theory for control in robotic systems.

The work is motivated by the desire to generate or augment dynamical models of robotic systems through data collection. In particular, it is of interest to synthesize model-based controllers using these data-driven models. Thus, the main contribution of this paper is the application of Koopman operator theory to the control of robotic systems. The Koopman operator is shown to have a linearizable data-driven model of the dynamical system that is amenable to model-based control methods. Closed-loop and open-loop controllers are then formulated using the proposed data-driven model. Furthermore, we explore the consequences of the specific choice of basis function as well as complexity order for swing up control of a simulated cart- and vertical take-off and landing (VTOL)-pendulum systems. Last, experiments using the Koopman operator using a Sphero SPRK robot are shown. We conclude the paper with recommendations for future work.

The organization of this paper is as follows. Section II gives an overview of the Koopman operator theory and its application to data-driven approximations of dynamical systems.  In addition, Sections III and IV explore the implementation of Koopman operator theory in simulation and experimentation, respectively. Conclusions are in Section~V.

\section{Koopman Operator}
An overview of Koopman operator theory is given in this section. For the purposes of this paper, we focus more on the practical implementation of the theory and omit much of the theoretical presentation. However, the interested reader can find a complete treatment of the Koopman operator in \cite{Koop2}.

To begin, consider a discrete-time dynamical system evolving as
\begin{equation} \label{dsystem}
x_{k+1} = F(x_k),
\end{equation}
where $x_k\in M$ is the, possibly unobserved, state of the system and $y_k \in \mathbb{C}$.  Furthermore, define an observation function
\begin{equation} \label{obeq}
y_k = g(x_k),
\end{equation}
where $g\in\mathbb{G}: M\rightarrow\mathbb{C}$ and $\mathbb{G}$ is a function space. For the purposes on this paper, we assume that $\mathbb{G}$ is the $L^2$ space. The Koopman operator, $\mathcal{K}: \mathbb{G}\rightarrow\mathbb{G}$, is defined as
\begin{equation} \label{Koopman1}
[\mathcal{K}g](x) = g(F(x)).
\end{equation}
Note that the Koopman operator maps elements in $\mathbb{G}$ to elements in $\mathbb{G}$. Therefore, it does not, as done by $F$, map system states to system states. Furthermore, note that (\ref{Koopman1}) can be written as
\begin{equation}\label{Koopman11}
[\mathcal{K}g](x_k) = g(F(x_k)) = g(x_{k+1}).
\end{equation}
Therefore, the Koopman operator propagates the output of the system forward.
Finally, the observable equation can be easily extended to the case where multiple observations are available, $g: M\rightarrow\mathbb{C}^K$.

The Koopman operator defined in (\ref{Koopman1}) is linear when $\mathbb{G}$ is a vector space. This property holds even if the considered discrete-time dynamical system is nonlinear. However, since the Koopman operator maps $\mathbb{G}$ to elements in $\mathbb{G}$ it is infinite dimensional. Therefore, a nonlinear dynamical system given by (\ref{dsystem}) can be equivalently described by a \emph{linear infinite dimensional operator}. From a practical standpoint, there is not much benefit from this infinite dimensional representation even if the operator could be defined for a specific system of interest. However, the Koopman operator can be approximated with a linear finite dimensional operator using data-driven approaches.

\subsection{Approximating a Koopman Operator}

In order to define an approximate Koopman operator the observation function (\ref{obeq}) is redefined as
\begin{equation}
y_k = g(x_k) = \Psi(x_k),
\end{equation}
where $\Psi(x)$ is a user-defined vector-valued function
\begin{equation}
\Psi(x) = [\psi_1(x), \psi_2(x), \dots, \psi_{N}(x)].
\end{equation}
Next, the relation described  by (\ref{Koopman11}) is now given as
\begin{equation} \label{Koopman2}
\Psi(x_{k+1}) = \Psi(x_k)K + r(x_k).
\end{equation}
where $K\in\mathbb{C}^{N \times N}$ and $r(x_k)$ is a residual (approximation error).
Note that the matrix $K$ advances $\Psi$ forward one time step. Next, it is assumed that the trajectory of the system has been collected such that
\begin{align} \label{datapoints}
X &=  [x_1, \dots, x_P]
\end{align}
where $P$ is the number of recorded data points.

The matrix $K$ can be computed in a number of ways. In this paper, we adopt the least-squares approach, described in \cite{williams2015data}, where $K$ is determined by minimizing
\begin{align}
J &= \frac{1}{2}\sum^{P-1}_{p=1}|r(x_p)|^2, \label{sumres}\\
&=\frac{1}{2}\sum^{P-1}_{p=1}|\Psi(x_{p+1}) - \Psi(x_p)K|^2.
\end{align}
Solving the least-squares problem yields
\begin{align}
K = G^\dagger A,
\end{align}
where $\dagger$ denotes the Moore--Penrose pseudoinverse and
\begin{align}
G &= \frac{1}{P}\sum^{P-1}_{p=1}\Psi(x_{p})^\textrm{T}\Psi(x_{p}), \\
A &=\frac{1}{P}\sum^{P-1}_{p=1}\Psi(x_{p})^\textrm{T}\Psi(x_{p+1}). \label{Amat}
\end{align}
Note that the computational burden of this approach grows as the dimension of $\Psi$ increases. The approach generally yields a better approximation as the dimension of $\Psi$ increases. Furthermore, the number of data points and their distribution across the state space will have a large effect on the computed $K$ matrix.

The definitions of (\ref{datapoints}-\ref{Amat}) can be generalized. The recorded data points need not come from a single trajectory nor be sequential \cite{williams2015data}. Multiple trajectories and trajectories with missing data points can be used. The only requirement is the sum of residuals given in (\ref{sumres}) be defined by consecutive states $(x_k,x_{k+1})$ spaced equally in time. Even this could be avoided by choosing another optimization to solve for $K$.

\subsection{Approximating Dynamical Systems}
For predicting dynamical systems, the approximation to the Koopman operator can be used to generate a data-driven model of a system by defining $\Psi$ as
\begin{equation}
\Psi(x) = [x^\textrm{T}, \psi_{n+1}(x), \dots, \psi_{N}(x)].
\end{equation}
Note that the state of the system, $x\in\mathbb{R}^n$, is now included in $\Psi(x)$.
Thus we can write the approximate dynamical equations of the considered system as
\begin{equation} \label{eom}
x_{k+1} \approx \hat{K}^\textrm{T}\Psi(x_k)^\textrm{T},
\end{equation}
where $\hat{K}^T \in\mathbb{R}^{n \times N}$ is the first $n$ columns of $K$.
Note that equation (\ref{eom}) simply propagates forward the quantities of interest (e.g. system states). Furthermore, in this work, $x_{k+1}$ is described as a linear combination of the system state, $x_k$, and the functions $\psi_i(x_k)$.

\section{Control Synthesis: Open- and Closed-loop Controllers}

In this section we formulate open- and closed-loop model-based controllers using the Koopman operator. It is first shown that for a differentiable choice of basis function $\Psi$, the Koopman operator has a linearization that can be computed for model-based control methods. Given the linearizable Koopman operator, a model-based optimal control problem is formulated for open- and closed-loop controllers.

\subsection{Koopman Operator Linearization}
By choosing a $\Psi$ that is differentiable, the Koopman operator approximation to the dynamical system can be linearized:
\begin{align}
x_{k+1} &\approx \hat{K}^\textrm{T}\frac{\partial\Psi}{\partial x}x_k \\
&\approx  A(x_k)x_k.
\end{align}

Control inputs are readily incorporated to the definition of $\Psi$ as an augmented state,
\begin{equation}
\Psi(x,u) = [x^\textrm{T}, u^\textrm{T},\psi_1(x,u), \psi_2(x,u), \dots, \psi_{N}(x,u)].
\end{equation}
This yields the approximate dynamical equations,
\begin{align}
x_{k+1} &\approx \hat{K}^\textrm{T}\Psi(x_k,u_k)^\textrm{T}
\end{align}
and the linearization of the approximate dynamical equations,
\begin{align}
x_{k+1} &\approx \hat{K}^\textrm{T}\frac{\partial \Psi}{\partial x}x_k + \hat{K}^\textrm{T}\frac{\partial \Psi}{\partial u}u_k \\
&\approx  A(x_k,u_k)x_k + B(x_k,u_k)u_k.
\end{align}
Note that \emph{linearizable equations of motion of a dynamical system can be computed solely from data}.

\subsection{Optimal Control Problem}

Control synthesis for trajectory optimization is generated for mobile robot dynamics of the form
\begin{equation} \label{eq:dynamics}
x_{k+1} = f(x_k, u_k),
\end{equation}
where $x \in \mathbb{R}^n$ is the state and $u \in \mathbb{R}^m$ is the control input. For a discrete system, we can solve for a trajectory that minimizes the objective defined as
\begin{equation} \label{eq:objective}
J = \sum_{k=0}^{N} \frac{1}{2}(x_k - \tilde{x}_k)^T\bold{P} (x_k - \tilde{x}_k) + \frac{1}{2}u_k^T \bold{R} u_k,
\end{equation}
where $\bold{P} \in \mathbb{R}^{n \times n}$ and $\bold{R} \in \mathbb{R}^{m \times m}$ are positive definite weight matrices on state and control and $\tilde{x}_k$ is the reference trajectory at time $k$. Note that the accuracy of the system model (\ref{eq:dynamics}) will largely determine the effectiveness of the synthesized optimal control.

\subsubsection*{Open-Loop}
Open-loop trajectory optimization precomputes the set of trajectory and control actions that minimize the objective function (\ref{eq:objective}) subject to the modeled dynamical constraints in (\ref{eq:dynamics}). Projection-based optimization \cite{hauser2002projection} is used in discrete time to generate the set of trajectory and control actions given an initial trajectory $x_k$ and control $u_k$ for $k \in \left[ 0, N \right] $. In the experiment, the projection-based optimization algorithm first generates the control actions based on the dynamical model and then at a fixed rate the command signals are sent via Bluetooth communication to the robot. Odometry data is collected only for post-processing and is not used to update the command signals.

\subsubsection*{Closed-Loop}
In the simulated and the experimental work, a discrete-time version of Sequential Action Control (SAC)~\cite{ansari2015sequential} is used with the Koopman operator to generate closed-loop optimal control calculations. However, any MPC technique can be used with the Koopman operator. Here, SAC operates by first forward simulating an open-loop trajectory for some horizon $N$ for a control-affine dynamical system given by
\begin{equation}
x_{k+1} = f(x_k, u_k) = g(x_k) + h(x_k)u_k .
\end{equation}
The sensitivity to a control injection for any given discrete time of the objective function is given as
\begin{equation}
\frac{dJ}{d\lambda_k} = \rho_k (f_2(k) - f_1(k))
\end{equation}
where
\begin{eqnarray}
f_1(k) &=& f(x_k, u_{0,k}), \\
f_2(k) &=& f(x_k, u_{k}^\star)
\end{eqnarray}
are the dynamics subject to the default control $u_{0,k}$ and derived control $u_k^\star$. The co-state variable $\rho_k \in \mathbb{R}^n$ is computed by backwards simulating the following discrete equation
\begin{equation}
\rho_{k-1} = \frac{\partial l_{k}}{\partial x} + \frac{\partial f_{k}}{\partial x}^T \rho_{k},
\end{equation}
where $l_k = \frac{1}{2}(x_k - \tilde{x}_k)^T \bold{P} (x_k - \tilde{x}_k) + \frac{1}{2}u_k^T \bold{R} u_k$ and $f_k = f(x_k, u_{0,k})$ for some default $u_{0,k}$ subject to $\rho_N = \vec{0}$. The optimal control $u^*_k$ is computed by first defining a secondary objective function as
\begin{equation} \label{secondaryObjective}
J_u = \sum_{k=0}^{N} \frac{1}{2} (\frac{dJ}{d\lambda_k} - \alpha_d)^2 +  \frac{1}{2} \Vert u_k^\star - u_{0,k} \Vert _\bold{R}^2.
\end{equation}
The objective (\ref{secondaryObjective}) is now convex in $u_k^*$ and has a minimizer when
\begin{equation}
u^*_k = (\Lambda + R^T)^{-1} h(x_k)^T \rho_k \alpha_d + u_{0,k},
\end{equation}
where $\Lambda = h(x_k)^T \rho_k \rho_k^T h(x_k)$.
Given the sequence of actions $u^\star_k$, it is then possible to calculate the time of control application $t_k^\star$ as
\begin{equation}
t_k^\star = \argmin \frac{dJ}{d\lambda_k}.
\end{equation}
The control duration in discrete time is found using an outward line search \cite{nocedal2006numerical} for a sufficient descent on the cost.

\section{Experiments Using Sphero SPRK}

\begin{figure}[h]
\centering
\includegraphics[scale=0.7]{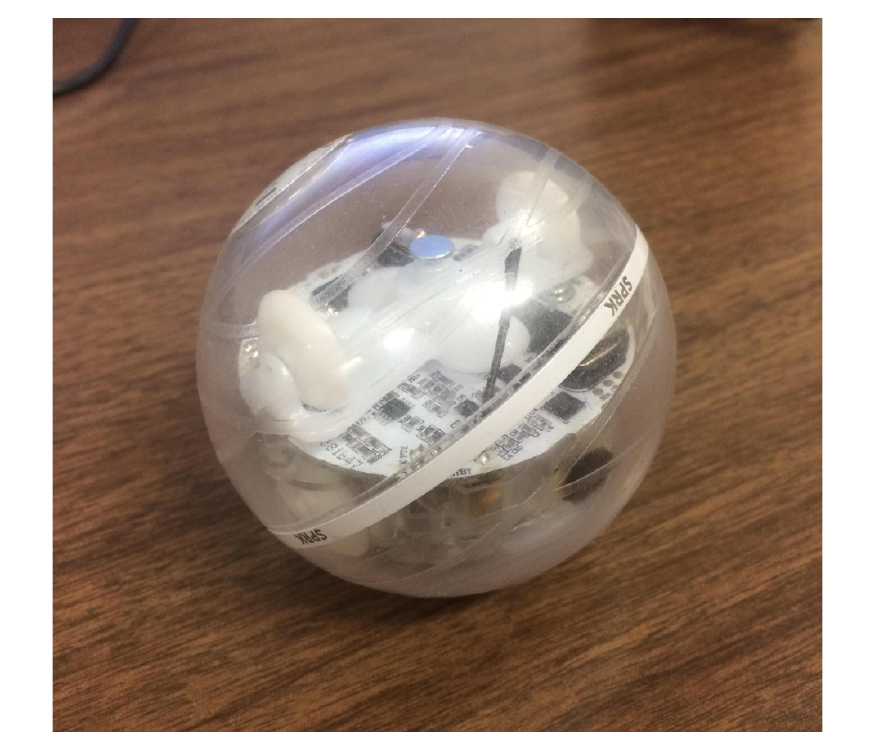}
\caption{ Sphero SPRK Robot is shown with its clear spherical casing revealing the underlying mechanism. The internal mechanism shifts the center of mass by rolling and rotating within the spherical enclosure, causing the SPRK to roll. RGB LEDs on the top of the SPRK are utilized to track the odometry of the robot through an Xbox Kinect with OpenCV and OpenKinect libraries for image processing and motion capture. ROS~\cite{ROS} is used to transmit and collect data at $20$Hz.}
\label{fig:SPRK}
\end{figure}

\begin{figure*}[th]
\includegraphics[width = \textwidth]{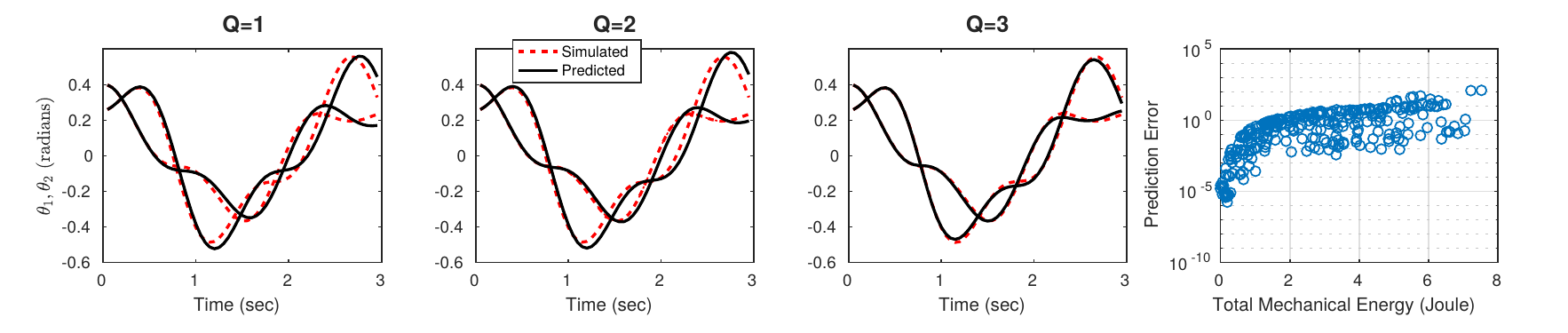}
\caption{ Simulated trajectories when the approximate Koopman operator was used to propagated the system's configuration. As the complexity $\Psi$ increases, so does the accuracy in prediction. 100 trials with uniformly random initial conditions were conducted to invesgate the relationship between accuracy and total mechanical energy. The prediction error tended to increase with total mechanical energy. }
\label{doublePen}
\end{figure*}

In this section, we describe the experimental set-up for use of the Sphero SPRK robot with model-based control algorithms that utilize a state-space model generated via the Koopman operator. In particular, we define data-driven closed- and open-loop model predictive controllers as well as motivate and explore the utility of Koopman operator for control of a robotic system.

In the experiments with the SPRK, trajectory optimization is run both in open-loop form and closed-loop feedback form. Here, the tracked states of the robot are position $x,y$ and velocity $\dot{x},\dot{y}$ and inputs to the robot are desired velocities $u_1, u_2$. The objective function parameters are defined as $\bold{P} = \text{diag}([60,60,0.1,0.1])$ and $\bold{R} = \text{diag}([20,20])$ and are maintained constant through both open-loop and closed-loop experiments. An additional set of experiments are done to show the use of the Koopman operator for control in a sand environment.

\subsection{SPRK}
The SPRK is a differential drive mobile robot enclosed in a spherical case. The dynamics of the SPRK are driven by the nonlinear coupling between the internal mechanism and the outer spherical encasing. In addition, proprietary underlying controllers govern how the command velocities are interpreted to low-level motors. The proprietary embedded software uses the on-board gyro-accelerometers to balance the robot upright while rolling. The caster wheels on top of the internal mechanism ensures constant contact of the lower wheels that are driven via two motors. The embedded software interfaces with heading and velocity (or $x-y$ velocity) command inputs sent via Bluetooth communication. A high fidelity model of the robot would include several internal states characterize the internal mechanism and controller. However, rather than seeking to approximated a high dimensional model, a reduced state model was sought.

Figure~\ref{fig:SPRK} shows a closer look at the SPRK robot. Odometry is collected using a Xbox Kinect with OpenCV~\cite{opencv_library} image processing. More details about odometry and motion capture are stated in the caption of Fig.~\ref{fig:SPRK}.

\subsection{SPRK Koopman Operator}
The representation of the system consists of the position of the robot $(x,y)$, its velocity $(\dot x, \dot y)$, and the commanded velocity $(u_x, u_y)$. Odometry data from the Kinect paired with recorded velocity commands are used to generate the approximate Koopman operator. The vector-valued functions used in this experiment are polynomial basis functions given as
\begin{align}
\Psi(x) &= [x,y,\dot x, \dot y, u_x, u_y, 1, \psi_1,\psi_2,\dots,\psi_M] \\
\psi_i(x) &= \dot x^{\alpha_{i}}\dot y^{\beta_{i}}
\end{align}
where $\alpha_i$, and $\beta_i$ are nonnegative integers, index $i$ tabulates all the combinations such that $\alpha_i + \beta_i\leq Q$ and $Q>1$ defines the largest allowed polynomial degree. We ignore higher order position dependence in the operator in order to prevent any possible overfitting of position-based external disturbances. The approximated Koopman operator was computed using data captured when the robot was operating at velocity under $1 \ m/s$ for the open-loop trails.

\section{Results}

\subsection{Simulation: Mechanical Energy}

In this section, the equations of motion of a double pendulum are approximated with the method described in Section II. The mass of both pendulums are 1 kilogram and the lengths of both are 1 meter. The mass of the pendulums are assumed to be concentrated at their ends. The system is conservative and subject to a gravitational field ($\textrm{9.81 m/s}^2$).

The state of the system, $x$, is described by the relative angles of the pendulums with respect to the vertical ($\theta_1$ and $\theta_2$) and their time derivatives ($\dot\theta_1$ and $\dot\theta_2$).
Data was collected by simulating the system multiple times with random initial conditions given by
\begin{align}
x_0 = [\mathcal{U}(-1,1)l_{\theta_1},\mathcal{U}(-1,1)l_{\theta_2}, \mathcal{U}(-1,1)l_{\dot\theta_1}, \mathcal{U}(-1,1)l_{\dot\theta_2}] \nonumber
\end{align}
where $\mathcal{U}(-1,1)$ is an uniformly distributed random variable with range $-1$ to $1$.
Furthermore, $l_{\theta_1}=l_{\theta_2}=\frac{\pi}{3}$ and $l_{\dot\theta_1}=l_{\dot\theta_2}=0.5$.
Therefore, the initial condition is uniformly distributed around the origin (and the stable equilibrium) and its range is defined by $L = [l_{\theta_1},l_{\theta_2}, l_{\dot\theta_1}, l_{\dot\theta_2}]$.
Any data point that fall outside of the range defined by $L$ was not used to approximate the Koopman operator.
Data collection occurred at 100 Hz and was stopped when $2,000$ data points were collected.

The vector-valued functions used in this numerical experiment are polynomial basis functions give as
\begin{align}
\Psi(x) &= [\theta_1,\theta_2,\dot\theta_1,\dot\theta_2, 1, \psi_1,\psi_2,\dots,\psi_M] \\
\psi_i(x) &= (\theta_1/l_{\theta_1})^{\alpha_{i}}(\theta_2/l_{\theta_2})^{\beta_{i}}(\dot\theta_1/l_{\dot\theta_1})^{\gamma_{i}}(\dot\theta_2/l_{\dot\theta_2})^{\delta_{i}}
\end{align}
where $\alpha_i, \beta_i,\gamma_i$, and $\delta_i$ are nonnegative integers, index $i$ tabulates all the combinations such that $\alpha_i + \beta_i + \gamma_i + \delta_i \leq Q$ and $Q>1$ defines the largest allowed polynomial degree.
Note that $-1\leq\psi_i\leq1$ when the state of the system is within the defined range.
The polynomial basis functions were scaled by the maximum expected value of the state to prevent numerical instability when higher order polynomials were utilized.

Figure \ref{doublePen} shows a simulated trajectory and the corresponding predicted trajectories when approximated Koopman operators were used to propagate the system's configuration.
As expected, the accuracy of the predicted trajectories are improved when Q is increased.
Figure \ref{doublePen} also shows how the accuracy of the predicted trajectories are dependent on the initial conditions.
The prediction error of a trajectory is computed as
\begin{align}
\frac{1}{N}\sum_i^N (x_{\textrm{sim},i} - x_{\textrm{K},i})^2
\end{align}
where $x_\textrm{sim}$ is the simulated trajectory, $x_\textrm{K}$ is the system's trajectory predicted by the approximated Koopman operator, and $N$ is the total run-time of the simulation. The prediction error tended to increase with total mechanical energy. Recall that the dynamics of a double pendulum are described by transcendental functions. Therefore, any approximation by polynomials of these dynamics will deteriorate as the relative angle increases in magnitude. However, when the relative angles are small (total mechanical energy is small) a polynomial approximation is accurate. As expected, selection of $\Psi$ plays a critical role in determining the quality of the computed Koopman operator.

\subsection{Simulation: Inversion and Stabilization of Pendulum Systems}

In this section, we describe the results of utilizing the Koopman operator for inverting a cart-pendulum system and a VTOL-pendulum system. In particular, this section overviews the effect that the choice of basis functions has on systems that have components in $SO(n)$ for $n > 1$.

For the cart-pendulum system, the Koopman states are given as

\begin{equation}
\Psi(x) = [\theta,x,\dot\theta,\dot x, u, 1, \psi_1,\psi_2,\dots,\psi_M]
\end{equation}
where we use
\begin{equation}
\psi_i(x) = \theta^{\alpha_{i}}x^{\beta_{i}}\dot\theta^{\gamma_{i}}\dot x^{\delta_{i}} u
\end{equation}
as the polynomial basis function set and compare with a Fourier basis function,
\begin{equation}
\psi_i(x) = \prod_{[x]_i} \prod_{\kappa_j}\cos([x]_i \kappa_j) \sin([x]_i \kappa_j) u,
\end{equation}
where $[x]_i$ is the $i^{th}$ state of the system and $\kappa_j$ is the $j^{th}$ basis order such that $\sum_j \kappa_j \le Q$.

\begin{figure*}[th]
\centering
\includegraphics[scale = 0.8]{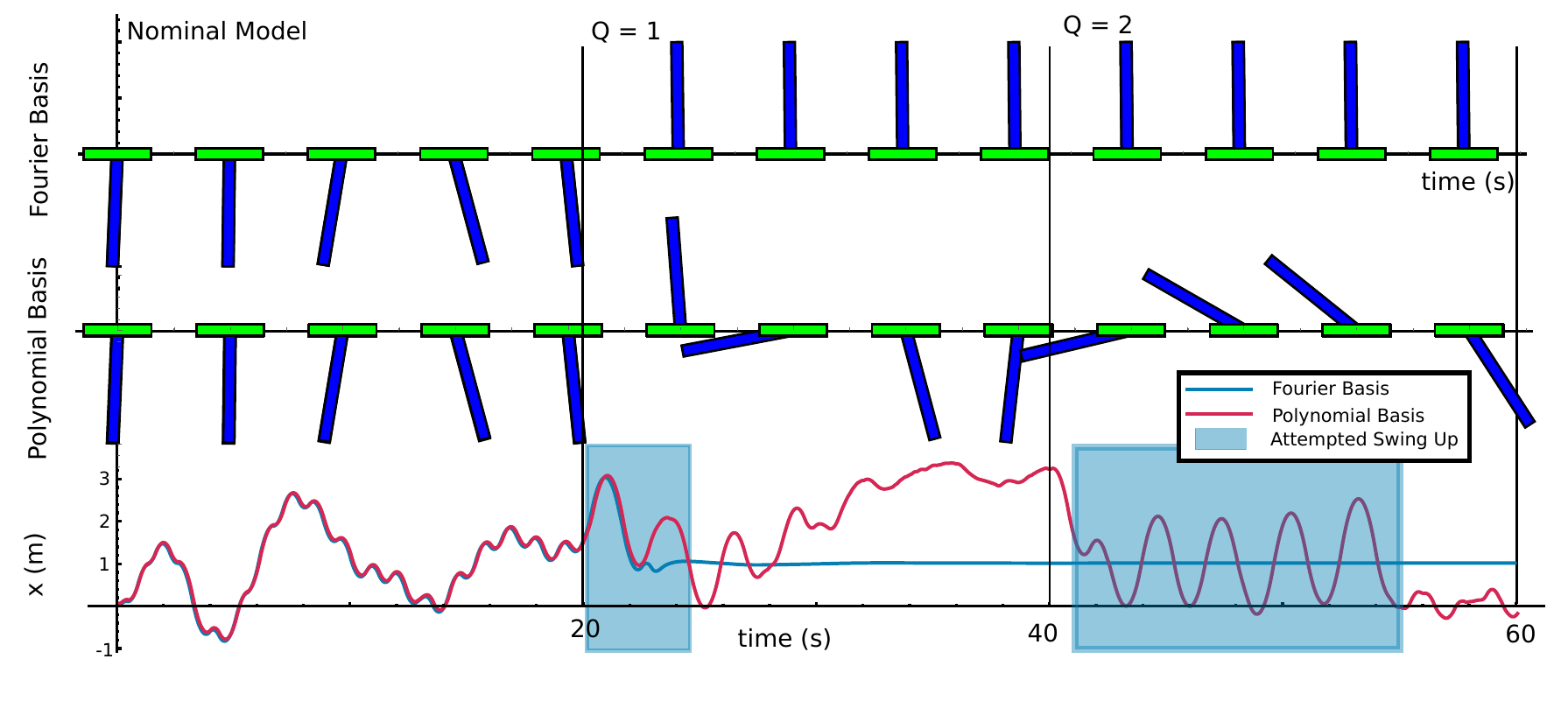}
\caption{ The progressive improvement in control as the Koopman operator increases the basis order of complexity $Q$ is shown. Each pendulum configuration is taken as a snapshot in time. Koopman operators with complexity $Q$ are trained on the initial first $20$ seconds with the nominal model. Note that because of the $SO(1)$ configuration of the pendulum, a Fourier basis of complexity $Q=1$ is sufficient to invert at stabilize the cart-pendulum. Adding a higher complexity $Q=2$ does not provide a different Koopman matrix (this does not necessarily hold true for non-simulated systems). It is interesting to note that as the complexity of the polynomial basis increases, so do the number of attempts at swinging up the cart-pendulum. Link to multimedia provided:  https://vimeo.com/219458009 . }
\label{CPSwingUp}
\end{figure*}

In this simulation, a nominal model given by
\begin{equation}
x_{k+1} = x_k +
\begin{bmatrix}
\dot\theta_k \\
\dot{x}_k \\
u\\
u
\end{bmatrix}  \delta t
\end{equation}
is utilized as an initial guess for the controller in order to boot-strap the data-driven process. Figure~\ref{CPSwingUp} presents the use of increasing complexity orders of a polynomial and Fourier basis function for the cart-pendulum system. Both test cases begin with the same initial condition and the same nominal model. At intervals of $20s$, a Koopman operator is computed with either the polynomial or Fourier basis functions using the initial $20s$ of data collected. Due to the existence of the pendulum on $SO(1)$, the Fourier basis function immediately generates a Koopman operator model that allows the controller to balance and stabilize the pendulum. Moreover, the use of the Fourier basis illustrates the concept that increasing complexity on the operator basis set is not always guaranteed to return an improved data-driven model. In particular, when $Q=2$, the Koopman operator matches the system model identically. As a result, any further additions in complexity using the Fourier basis for this system is not beneficial (this is not always the case if the system has higher order dependencies).  In contrast, the polynomial basis function does show improvement as complexity is increased. Although it would require an infinite set of polynomials to approximate a cosine or sine function, the controller using this operator model provides the desired energy pumping cart motion that is commonly witnessed in inverting a pendulum.

Simulated examples are further investigated with the use of a vertical take-off and landing (VTOL) pendulum system \cite{luukkonen2011modelling}. For this example, the problem of inverting the pendulum attached to a VTOL is slightly modified. Specifically, it is assumed that a well known model of the VTOL exists, but the interaction between the VTOL and the pendulum remains unknown. Thus, the goal of this simulated example is to generate a Koopman operator that describes the interaction of the VTOL on the pendulum.

\begin{figure*}[th]
\centering
\includegraphics[scale = 0.8]{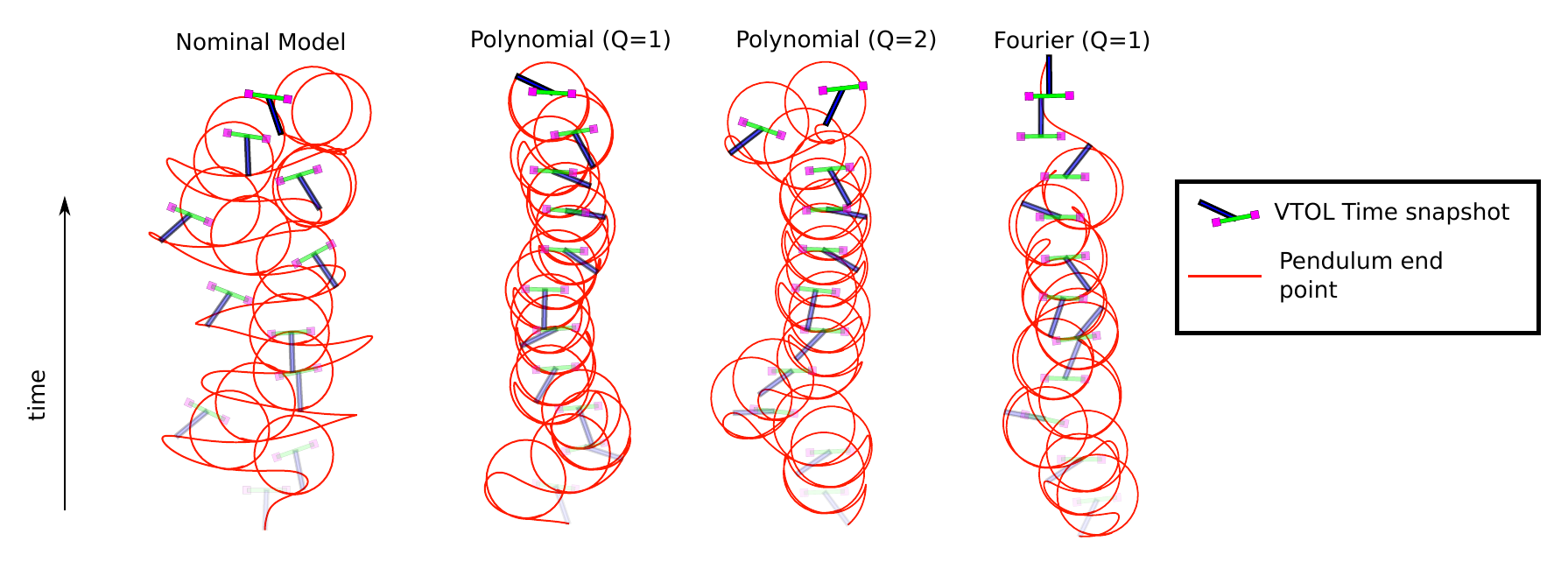}
\caption{ Each Koopman operator is trained on the residual modeling error of $20$ seconds attempted pendulum inversion using the nominal model. As the order of the polynomial basis increases from $1 \to 2$, the number of swing up attempts also increases. Notably, a first order Fourier basis captures the necessary features that allow the controller to invert and stabilize the pendulum. Link to multimedia provided:  https://vimeo.com/219458009 .}
\label{vtolSwingUp}
\end{figure*}

In this example, the Koopman operator is redefined as an augmentation to a dynamical system
\begin{equation} \label{augKEOM}
x_{k+1} = f(x_k, u_k) + \tilde{K}^{T} \Psi(x_k, u_k)^T.
\end{equation}
By subtracting the current nominal model of the system $f(x_k, u_k)$ from both side in equation (\ref{augKEOM}) and treating $x_{k+1}$ as the measurement of state, we can define the following as a nonlinear process that can be used to generate a Koopman operator:
\begin{equation}
\Psi(x_{k+1}) = x_{k+1} - f(x_k, u_k) = \tilde{K}^T \Psi(x_k, u_k)^T.
\end{equation}
Given the previous cart-pendulum result, we see that the interaction between the VTOL and the pendulum can be captured solely via a vast set of basis functions across the state of the VTOL-pendulum system. In Fig.~\ref{vtolSwingUp}, the VTOL is shown attempting to invert and balance the pendulum attached with the use of the Koopman operator. Each sequential Koopman operator with increasing complexity is generated from the first $20$ seconds worth of data. Originating from the nominal model, it can be seen that the swinging behavior captures a portion of the energy pumping maneuvers required to invert the pendulum. As the Koopman basis order increases, so does the refinement in control authority. When $Q=2$ for the polynomial basis, it can be seen that swing up attempts are more successful. Once the Koopman operator generated from the Fourier basis functions is used, the controller generates the appropriate control strategy to swing up and invert the pendulum.

In the following section, our discussion on the use of the Koopman operator is extended to control of a Sphero SPRK robot in a reduced state setting.

\subsection{SPRK Experiments}

\subsubsection{Open-Loop Trajectory Optimization}

\begin{figure*}[!ht]
\includegraphics[width=\textwidth]{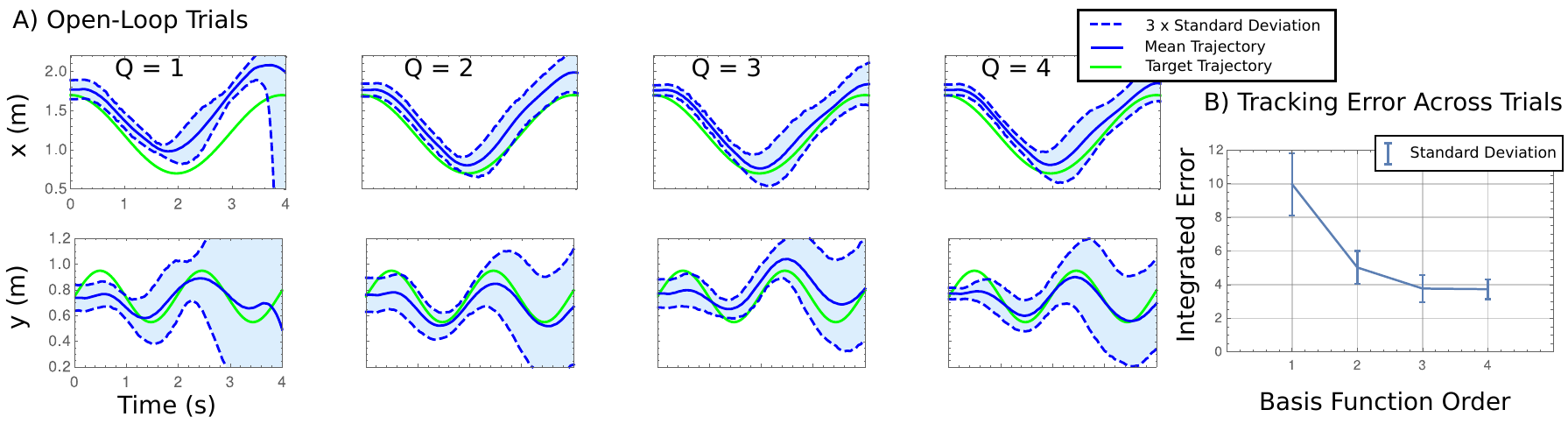}
\caption{ Here we show reference tracking using open-loop trajectory optimization. The reference trajectory was made sufficiently aggressive to excite the system's internal nonlinearities that cannot be captured completely by the minimal state representation.
Respective integrated tracking errors are shown to decrease with an increase in $Q$.
This suggests that the approximate Koopman operator better represents the dynamics of the system with increasing complexity of $\Psi$. }
\label{fig:openloop}
\end{figure*}

Figure~\ref{fig:openloop} shows trajectories generated using the open-loop controller with varying $Q$.
The reference trajectory is given as
\begin{equation} \label{fig8}
\begin{bmatrix}
\tilde{x}\\
\tilde{y}\\
\dot{\tilde{x}}\\
\dot{\tilde{y}}\\
\end{bmatrix}
 =
\begin{bmatrix}
r \cos(v t) \\
r \sin(2v t) \\
-r v \sin(v t ) \\
2rv\cos(2v t)
\end{bmatrix}.
\end{equation}
where  $r=0.5$ and $v = 1.3$.
The reference trajectory was made sufficiently aggressive to excite the system's internal nonlinearities.

As expected, the system improves in performance when tracking the reference trajectory with increasing $Q$. In particular, as $Q$ goes from $1$ to $2$, less drift in the resulting open-loop trajectory is visually noted at the end of the path. As $Q$ is further increased, more complexity is added to the description of the SPRK via the Koopman operator which in turn reduces drift and improves the tracking performance. Furthermore, the standard deviation of tracking error across trials is shown to reduce as $Q$ is increased. This implies both consistency in the behavior of the robot subject to the controller. Therefore, it can be concluded that the approximated Koopman operator is better able to represent the dynamics of the system by increasing the complexity of $\Psi$.

\subsubsection{Closed-Loop Trajectory Tracking}

\begin{figure*}[th]
\centering
\includegraphics[scale=1]{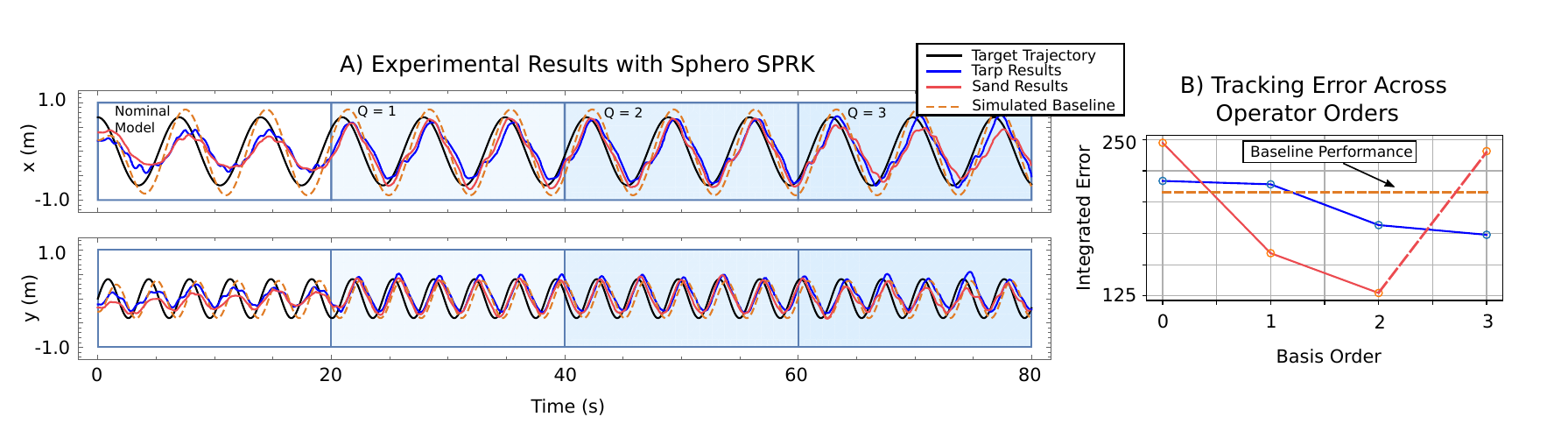}
\caption{ Here, we show closed-loop model-based control using sequentially increasing basis complexity, $Q$ in the Koopman operator. Two examples using the SPRK robot are run on a tarp and on sand. A baseline simulated example is provided to show the best-case performance of the controller subject to the nominal model used. As the complexity of the operator's basis function is increased, so does the performance of the tracking. Note that in B), the $3rd$ order operator used in sand (shown as the dashed red line) did not have a sparse enough set of data to provide a stable model, although it performed better than the nominal model. Link to multimedia provided:  https://vimeo.com/219458009 .}
\label{fig:closed_loop_experimental}
\end{figure*}

Figure~\ref{fig:closed_loop_experimental} shows the experimental results for trajectory tracking on a tarp and sand terrain using closed-loop model-based controllers with the Koopman operator. The optimal control signal was updated at 20 Hz and the reference trajectory was given by equation (\ref{fig8}) where $r$ is split into two components, $r_x=0.7$ and $r_y=0.4$, with $v=0.9$. The nominal linear model is given by
\begin{equation}
x_{k+1} = A x_k + B u_k,
\end{equation}
where $A$ and $B$ are defined as a fully controllable double integrator system.

The effectiveness of the closed-loop controller is benchmarked by comparing the model generated from the Koopman operator to that of a simulated example of the controller knowing the true system model (Fig.~\ref{fig:closed_loop_experimental}). Using only the first $20$ seconds worth of data from the nominal model controller, we can see in Fig.~\ref{fig:closed_loop_experimental} A) that as the operator increases in complexity, so shows the performance of the controller relative to the benchmark test. Specifically, Fig.~\ref{fig:closed_loop_experimental} B) shows the tracking error for experimental trials with increasing complexity of the Koopman operator. Notably, when $Q=3$ in sand, the Koopman operator did not have a sparse enough data set that spans the higher order terms in the operator. This can be fixed by collecting more data that spans the robot's operating region.

Here, the nonlinear dynamics driven by the internal mechanism become more apparent as the order of the operator is increased. In particular, equation (\ref{kEQ}) provides some insight into the output of the data-driven model of the Koopman operator for the update equation of the SPRK's velocity subject to control inputs. Because the effect of the internal mechanism's configuration (typically described on $SO(3)$) cannot be linearly approximated, the Koopman operator begins to approximate a Taylor expansion (\ref{kEQ}).
Therefore, the Koopman operator captures the inherent nonlinearities that are utilized by the model-based controller with respect to the terrain. However, achieving a representation that performs consistently across all operating terrains seems infeasible with such limited information, without extra structure on the Koopman operator, such as global Lie group structure or mechanical properties (e.g. symmetries). 

\begin{figure*}[!t]
\normalsize
\setcounter{MYtempeqncnt}{\value{equation}}
\setcounter{equation}{5}
\begin{equation}
\label{kEQ}
\begin{bmatrix}
\dot{x}_{k+1} \\
\dot{y}_{k+1}
\end{bmatrix}
=
\begin{bmatrix}
0.08 x_k - 0.35 y_k + 0.76 \dot{x}_k  +  0.21 \dot{y}_k  +  1.06 u_1 - 0.17 u_2 - 0.05 \dot{x}_k^2  - 0.19 \dot{y}_k^2   - 1.09 \dot{x}_k \dot{y}_k^2 - 0.71 \dot{y}_k \dot{x}_k^2   + 0.40 \\
  -0.06 x_k   + 0.16 y_k +   0.17 \dot{x}_k +  0.87 \dot{y}_k   -0.38 u_1 +   0.57 u_2  - 0.20 \dot{x}_k^2  - 0.52 \dot{y}_k^2 -0.45 \dot{x}_k \dot{y}_k^2 -3.17 \dot{y}_k \dot{x}_k^2  -0.04
\end{bmatrix}
\end{equation}
\setcounter{equation}{\value{MYtempeqncnt}}
\vspace*{4pt}
\end{figure*}

\vspace{-0.2in}
\section{Conclusion}

We present Koopman operator theory and focus on the practical implementation of the theory for model-based control.
We derive a linearizable data-driven model using the Koopman operator.
Closed-loop and open-loop controllers were formulated using the proposed data-driven model.
The open-loop experiments reveal the Koopman operator improves performance as the complexity of the basis increases.
Closed-loop experiments reveal the Koopman operator is able to capture the nonlinear dynamics of simulated examples with the cart- and VTOL-pendulum and the SPRK robot.

Future research directions include an in-depth analysis of the choice of basis for dynamical system with distinct structure (e.g. conservative systems, mechanical systems, etc.).
The relationship between available states and the accuracy of the approximate Koopman operator needs rigorous stability analysis.
Moreover, numerical stability analysis and algorithmic optimization is another possible research avenue.

\bibliographystyle{IEEEtran/IEEEtran}
\bibliography{IEEEabrv,RSS2017}

\end{document}